\documentclass{appolb}
\usepackage{amsmath,amssymb,graphicx,epsfig,pst-all,srcltx}
\DeclareMathOperator{\sgn}{sgn}

\begin{document}
\pagestyle{plain}
\title{Decisional Processes with Boolean Neural
Network: the Emergence of Mental Schemes
\thanks{Presented at Summer Solstice 2009 International Conference
on Discrete Models of Complex Systems Gdansk, Poland, 22-24 June
2009}
}

\author
{
Graziano Barnabei
\address{Dept. of Systems and Computer Science, University of Florence, via S. Marta 3. 50139 Firenze, Italy;
         CSDC, Via G. Sansone, 1. 50019 Sesto F.no, Firenze, Italy. email: grbarnabei@gmail.com;}
\\[.4cm]
 Franco Bagnoli
\address{Dept. of Energy, University of Florence, via S. Marta 3. 50139 Firenze, Italy;
         CSDC and INFN, sez. Firenze. email: franco.bagnoli@unifi.it;}
\\[.4cm] Ciro Conversano, Elena Lensi
\address{Dept. of Psychiatry, Neurobiology, Pharmacology and Biotechnology,
        University of Pisa, Via Roma, 67. 56126 Pisa, Italy.}
}

\maketitle

\begin{abstract}

Human decisional processes result from the employment of selected
quantities of relevant information, generally synthesized from
environmental incoming data and stored memories. Their main goal is
the production of an appropriate and adaptive response to a
cognitive or behavioral task. Different strategies of response
production can be adopted, among which haphazard trials, formation
of mental schemes and heuristics. In this paper, we propose a model
of Boolean neural network that incorporates these strategies by
recurring to global optimization strategies during the learning
session. The model characterizes as well the passage from an
unstructured/chaotic attractor neural network typical of data-driven
processes to a faster one, forward-only and representative of
schema-driven processes. Moreover, a simplified version of the Iowa
Gambling Task (IGT) is introduced in order to test the model. Our
results match with experimental data and point out some relevant
knowledge coming from psychological domain.

\end{abstract}

\section{Introduction}

Humans usually categorize incoming information into stable concepts
which can be upgraded, related and nested one into another. The
characteristics of information are analyzed and classified into
(i.e. they activate) existing concepts but, whenever they would
represent a novelty, they will induce the formation of a new concept
or the upgrade of the existing ones. This adaptive modality of
knowledge organization makes cognitive system able to classify,
store and employ at best incoming information, in order to solve the
eventual cognitive demand during next steps of processing.
Subsequently, human cognitive or behavioral responses to a given set
of inputs are built following several and different decisional
strategies. The role of context, the kind of information and past
experiences are central for the choice of what kind of decision
making will be made. In general, we can take into consideration at
least five strategies of output production:

\textbf{1) Reflexive responses}: direct associations of inputs with
an output pattern. They require no attentional resources and are out
of possible controls. Typical examples are reflexive genetically
implemented motor responses (e.g. the evade reflex) and associative
behaviors (e.g. the Pavlov's dog salivation reflex).

\textbf{2) Automatic processes}: standardized quick responses
associated to a frequent activation of simple concepts while the
behavioral relevance of the input/event is under an ``alert
red-line'' (i.e. it requires a behavioral response but not a direct
attentional monitoring, for instance during the vehicle driving).

\textbf{3) Routine processes}: processes triggered by several
related concepts or complex events sufficiently frequent to
constitute a stereotypical routine. Routines can be solved by a
$script$ \cite{1,2} and need the emergence of \textit{mental
schemes} \cite{2,3,4}, namely representations of complex concepts or
events easily connectable to a fast cognitive or behavioral
response. Note that even if the strategy requires an attentive
control, it doesn't involve the same set of cognitive ability needed
during a problem solving task, like the resolution of a syllogism or
the Wason selection task \cite{5}.

\textbf{4) Reasoning}: higher cognitive strategy of understanding
and production, mainly used during the problem solving. Given a set
of premises, humans seem to employ rules like those involved in
formal logic \cite{6}, which establish the correct formal solution.
The propositional reasoning makes no distinctions about the contents
of a statement, but deals only with its syntactical structure.
Unfortunately, human judgements are sometimes very far from correct
formal solutions. For these reasons, a theory of \textit{mental
models} \cite{7,8} has been proposed, which claims that
hypothetical-deductive reasoning have three stages of thought: an
understanding of the premises which leads up to model construction,
a formulation of provisional conclusions, a revision procedure that
verifies if other models are possible. Errors occur because of
working memory limitedness: the bigger is the number of models that
we have to menage, the harder the problem becomes. So, errors are
conclusions not rigorously verified.

\textbf{5) Heuristic behavior}: \textit{modus operandi} typical of
situations in which there is either a lack of information, or
different mental schemes run in conflict, or there is no time to
reasoning, or the task is too difficult. In these cases, individuals
adopt strategies more similar to an attempt rather than to the
formal solution. Note that, in some cases, the use of heuristics is
mandatory and constitutes a cognitive bias\footnote{While the
abstract version of the Wason selection task leads to a correct
performance of 3.9\%, the concrete one leads to a correct
performance of 91\%. These results are interpreted recurring to the
availability heuristic applied on past experiences.}. Most important
heuristics in psychology are anchoring/adjustment, availability, and
representativeness \cite{10}.

While strategies 1-4 belong to a hierarchy of use and exploitation
of cognitive resources, heuristics take place only after strategy 3,
and if there are no conditions to apply the other ones or their
application fails. Finally, it is possible to bring back the
aforementioned considerations into the simple cognitive model
reported in Fig.~1.
\begin{figure}
\begin{center}
\includegraphics[width=0.8 \textwidth, bb= 5 5 675 515]{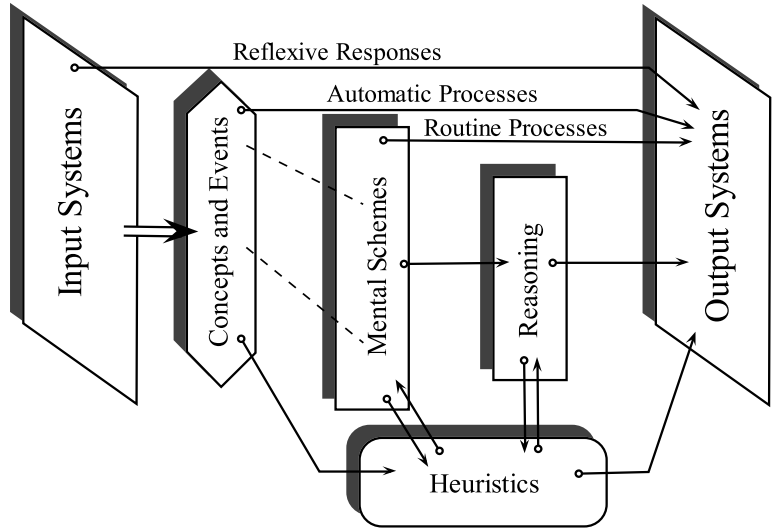}
\end{center}
\caption{The cognitive model. Flowchart symbols as defined in
\cite{14}}
\end{figure}
Neglecting earlier input systems and concept stages, our main
purpose is to formalize into a neural model the cognitive constructs
of mental scheme and heuristic, showing how these can modify the
production of a response. Besides we will propose a simplified
version of the IGT \cite{11,12} in order to test our model.

The paper is organized as follow. In the next section we introduce
the neural model, fourth section is devoted to model fitting, and in
the final sections we show the main results of our simulations and
discuss about future perspectives.

\section{The Model}
The model assumes that cognitive activities during a task resolution
can be represented as a Boolean neural network, whose nodes do not
necessarily correspond to single biological neurons but rather to
organized sets of neurons, named functional areas. We choose this
formalization in order to remain within the framework of neural
domain.

The basic computational entities, namely the formal neurons, are
described by the following parameters:

\begin {enumerate}
    \item $\sigma$: the internal and external state of activation
    $\in$ \{-1, +1\}
    \item $b$: the threshold $\in$ [0, 1]
    \item $c$: the connectivity degree, i.e. the number of afferent synapses
    of the neuron
\end {enumerate}

The $N$ bipolar neurons are linked by connections, named synapses,
each bearing a weight $\in$ \{-1, 0, +1\}\footnote{The case 0-weight
corresponds to the absence of link. We don't allow auto-synapses, or
self-recurring links.}. At time $t$, the $i$-th node computes
incoming signals from afferent neurons and, at time $t+1$, produce a
signal, i.e. fires, according to the following update law:

\begin{equation}
    \sigma^{t+1}_i = \sgn \left( \sum^{N}_{j=1} \frac {w_{ij} \cdot\
    \sigma^{t}_i}{c_i} - b_i \right) \label{dynamics}
\end{equation}

where sgn($x$) returns the sign of real number $x$, $w_{ij}$ is the
incoming weighted synapse of $i$-th neuron from the $j$-th one, and
$c_i=\sum_{j} |W_{ij}|$. This formalization makes the adopted formal
neuron similar to that of McCulloch \& Pitts \cite{9}.

\textbf{Dynamics: search of an asymptotic
configuration}\label{quick}. By generating an arbitrary
$\vec{\sigma}$ and $\vec{b}$, and a connection structure $W$ with
entries $w_{ij}$  uniformly distributed in \{-1, 0, +1\}, we are
able to define the starting configuration $\zeta^{t_0}$, composed by
$(W, \vec{b}, \vec{\sigma}^{t_0})$\footnote{We have just introduced
the vector notation for the matrix of synaptic weights $W =
(w_{ij})_{ij}$, the state vector $\vec{\sigma}^{t} = (\sigma^t_1,
\dots, \sigma^t_N)$, the threshold vector $\vec{b} = (b_1, \dots,
b_N)$ and the connectivity vector $\vec{c} = (c_1, \dots, c_N)$.},
i.e. the initial condition of the dynamics. Neurons are
synchronously updated by applying iteratively Eq.~(\ref{dynamics})
for a sufficiently large time $t_{meas}$, the maximum
\textit{convergence time} allowed\footnote{In principle the longest
convergence time $t_{c}$ should be $2^N$, the maximal periodical
orbit of a finite size discrete system of $N$ unities having 2
possible values. But for reasons of feasibility of simulations,
fixing a reasonable $t_{max}$, $t_{meas}$ will correspond to $\min
\{t_{c}, t_{max}\}$.}. If $W$ is asymmetric and according to its
asymmetry degree, a periodical orbit of length $l$ is reached after
a transient $\tau$, giving the asymptotic configuration
$\zeta^{t_{meas}}$, composed by $(W, \vec{b},
\vec{\sigma}^{t_{meas}})$. In the following, all the procedures will
make use of this concept of asymptotic configuration and its
eventual distance from the correct behavior. This is motivated from
the assumption that only stable asymptotical configurations can
account for stability and invariance of response typical of human
cognitive processes. In principle, $t_{c}$ can be viewed as the time
needed by the cognitive elaboration.

\textbf{Training phase: a problem of global optimization}. We choose
a subset of $n$ Boolean functions as possible instances of the
training problem. For each $\gamma$-th function we select
$N^{\gamma}_{I}$ input neurons from which the remaining ones take
entries for the update dynamics and just one, that is
$N^{\gamma}_{O}=1$, output neuron from which, after the attainment
of a $\zeta^{t_{meas}}$, we will measure the computed output.

Moreover, for each $\gamma$-th function we generate the
corresponding training set $\varepsilon^{\gamma}$, for which all
possible examples are given by coupling one of the inputs
$\xi^{\gamma, inp}_{\mu}$ with the respective wanted output
$\xi^{\gamma, out}_{\mu}$, for $\mu=1,...,P^{\gamma}$. The
presentation of the entire training set $\varepsilon$ is
consequently just one training epoch\footnote{Consequently, given
$n$ functions to learn, the input vector $I$ will have $N_{I} =
\displaystyle \sum^{n}_{\gamma=1} N^{\gamma}_{I}$ elements, the
output vector $O$ will have $N_{O} = \displaystyle
\sum^{n}_{\gamma=1} N^{\gamma}_{O} = n$ elements, and the size of
$\varepsilon$ will be $P=\displaystyle \sum^{n}_{\gamma=1}
P^{\gamma}$, where $P^{\gamma}=2^{N^{\gamma}_{I}}$. Notably, the
input vector $I$ cannot change entries during dynamical update
$\Rightarrow$ $T$ becomes $2^{N-N_{I}}$. We notice that, since we do
not need to test our network on the complementary subset of
$\varepsilon^{\gamma}$, we can freely present, for each $\gamma$-th
instance of the problem, the entire correspondent training set
$\varepsilon^{\gamma}$ during the training phase.}. At the end of
each epoch, the error signal $E(\zeta^{t_{meas}})$ will be:

\begin{equation}
E=
\begin{cases}
\displaystyle \frac {1}{4 P l_{w}} \sum^{N}_{i=N-N_{O}+1}
\sum^{P}_{\mu=1} \sum^{l_{w}}_{j=1} \left(S^{out}_{i, \mu} -
\xi^{out}_{i, \mu}\right)^2 + \frac {\tau_{w}-1}{10} & \text{if
$t_{c} \leq t_{max}$}
\\
10 \cdot N_{O} & \text{otherwise}
\end{cases}
\end{equation}
where $S^{out}_{i, \mu}$ and $\xi^{out}_{i, \mu}$ are respectively
the computed and the wanted output for the $\mu$-th pattern of
$\varepsilon$\footnote{We remark that the first term in first case
of Eq.~(2) is just the normalized average Hamming distance
generalized for the entire orbital length.}. The addiction of a term
for transient will justify the passage from a computationally slow
attractor neural network, prototypical of data-driven processes, to
a faster and oriented one representative of schema-driven processes.
Each asymptotic configuration having $E=0$, is one of the possible
configurations able to solve the $n$ incoming instances of the
problem and therefore the presented task.

As there aren't clues as the way $\zeta^{t_{meas}}$ have to be
modified, we choose to proceed by \textit{trial and error}.
Consequently, the learning problem can be associated with a problem
of global optimization, where $E$ is the object function to be
minimized. The choice of the optimization strategy will produce
different behaviors of the network during the learning phase and
consequently, we will be able to associate them to different
cognitive strategies of task resolution.

\textbf{Haphazard trials.} No optimization strategy is applied.
Starting from an arbitrary generated condition, a series of local
perturbations are produced, by modifying just one entry either in
$W$ or $b$ selected at random with uniformly distributed
probability. For each perturbation, the corresponding
$E(\zeta^{t_{meas}})$ is calculated. The resulting trend is a random
walk of $E$. This \textit{non strategic} behavior is directly
affected by the $N$ involved functional areas, namely by the size of
the solution space. Consequently this strategy can need a very long
time to reach the solution of the problem and to produce the correct
response to the task.

\textbf{Emergence of mental schemes.} The optimized asymptotical
configuration must be able to store the $n$ presented instances as
mental schemes, which future activation will produce a fast and
cognitively cheap response. We choose to formalize the emergence of
mental schemes as a simulated annealing procedure with geometrical
cooling ratio $cl$, fixed once for all at $0.6$, and by using $E$ as
energy. By starting from the arbitrary generated $\zeta^{t_{meas}}$,
a local perturbation is produced with same modalities of the
\textit{haphazard trials}. The resulting $\zeta^{'t_{meas}}$ differs
in $E$ of a value $\Delta E$ from the previous one. For the
acceptation of the new configuration, we refer to the Metropolis
algorithm. The new configuration is kept with probability:
\begin{equation}
p=
\begin{cases}
1 & \text {if $\Delta E<0$}
\\
\exp (-\Delta E/T) & \text {otherwise}
\end{cases}
\end{equation}
where $T$ will modulate the cooling schedule\footnote{The starting
$T_0$ is fixed once for all at 5. Each 10 epochs we sample the
acceptance probability $ap$. If $ap=0.5 \pm 0.2$ then $T_{k+1} = cl
\cdot T_k$.}. The perturbation procedure continues until
$E(\zeta^{t_{meas}})=0$ for a reasonable number of epochs. At this
point the system have inferred and stored the $n$ instances. When a
future behavioral situation will pose the same set of input, the
stable reached configuration will be reactivated and, having
respectively $\tau$ and $E$ equals to zero and $l$ equal to 1, it
will produce one of the fastest and correct responses admissible by
the task.

\section {Results}
Figure 2 shows how the attainment of a configuration having $E=0$
can depend on the size of the feasible region, in our model function
of $N$. During a series of local perturbations all accepted, the
greater is $N$ the more difficult becomes the search of a
configuration able to solve the task.
\begin{figure} [h]
\begin{center}
\begin{tabular}{cc}
a&b \\ \\ \hspace{-.7truecm}
\includegraphics[width=0.51 \textwidth, bb= -125 175 721 650]{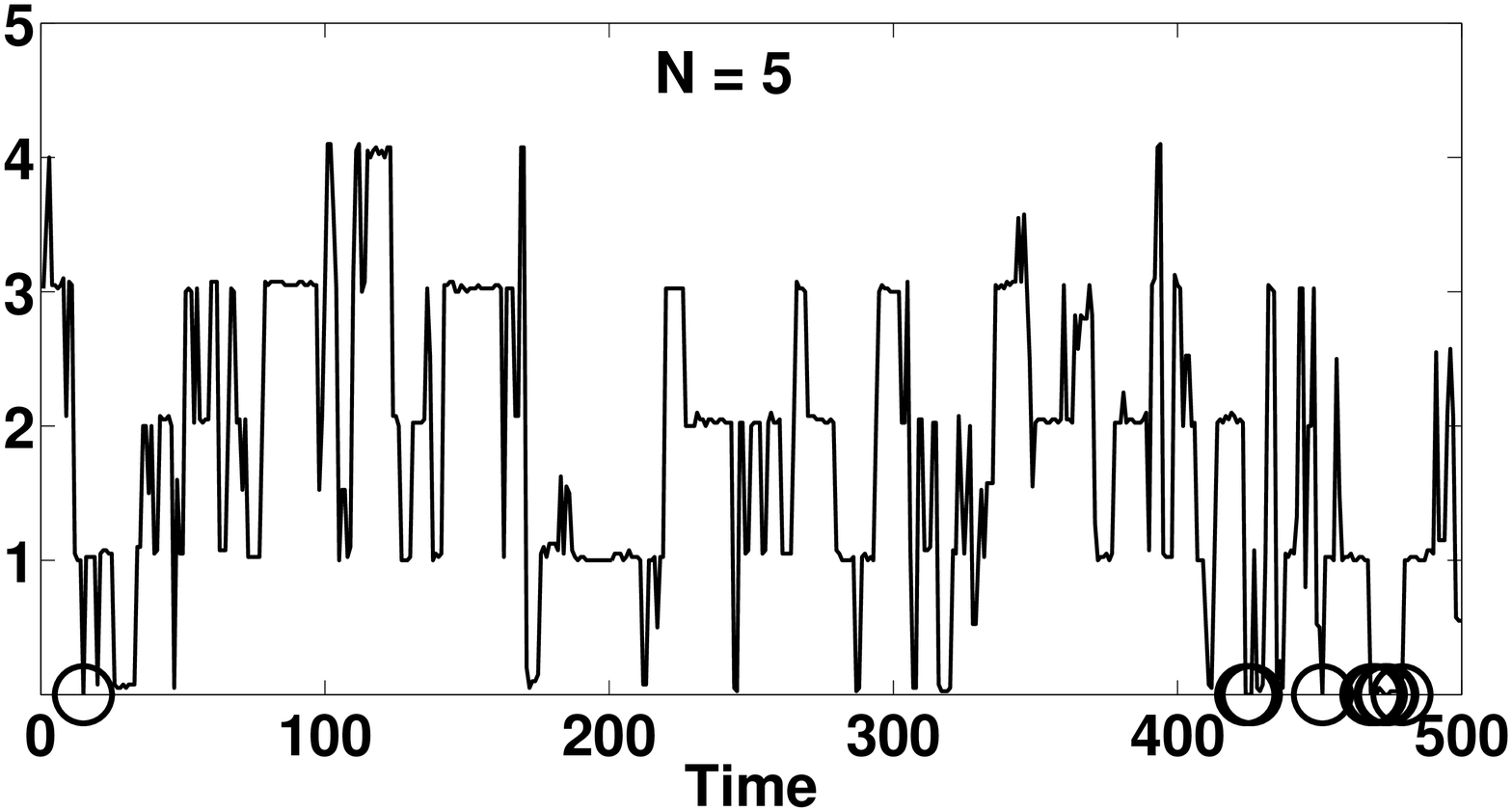}
&
\includegraphics[width=0.505 \textwidth, bb= -125 175 721 650]{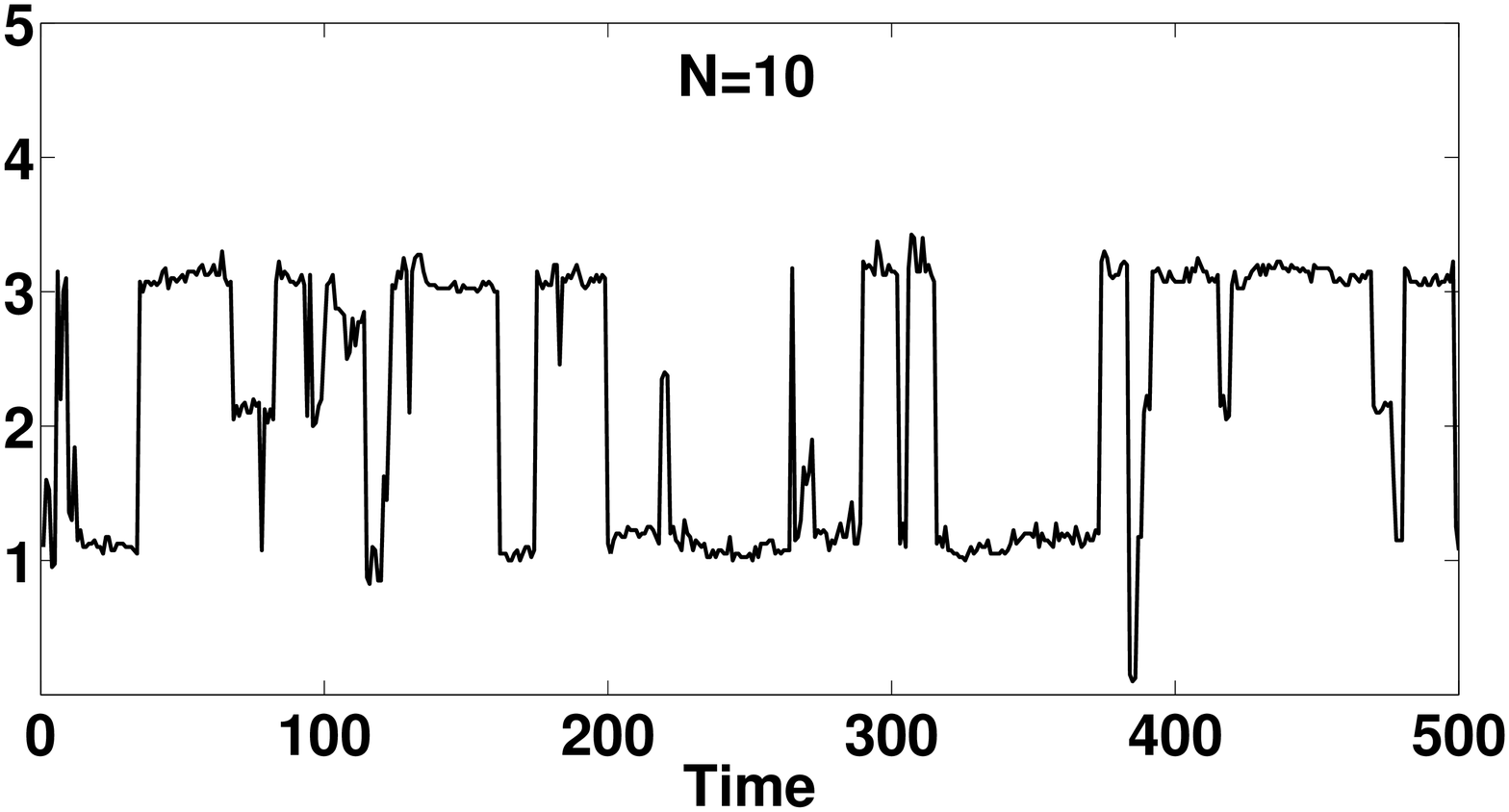}
\end{tabular}
\end{center}
\vspace{-0.7truecm} \caption{Error signal during perturbation phase.
Each time step is just one local perturbation. The circles marks the
occasional configurations with $E=0$. a) $N=5$; b) $N=10$.}
\end{figure}
Anyhow, this dependency is also found  by applying the algorithms of
global optimization during the learning phase.

Results of the optimization procedure are presented in Fig.~3.
Comparing Fig.~3a with Fig.~2b, it is clear what happens to the slow
dynamics during optimization. Having at the beginning a large
temperature $T$, all moves can be accepted in spite of their
respective error $E$, allowing the passage among basins. By
decreasing $T$, only moves that decrease the error $E$ begin to be
accepted, see Eq.~(4), causing a more exhaustive exploration of the
small-$E$ of the basin up to the reaching of the global minimum. The
dependence of $E$ from $n$ and $N$ shown in Fig.~3b,c can be easily
reported to the task difficulty, typically correlated with the
number of instances of the learning problem and the number of
involved functional areas.
\begin{figure} [h]
\begin{center}
\begin{tabular}{ccc}
a&b&c \\ \\ \hspace{-.7truecm}
\includegraphics[width=0.325 \textwidth, bb= -125 175 721 650]{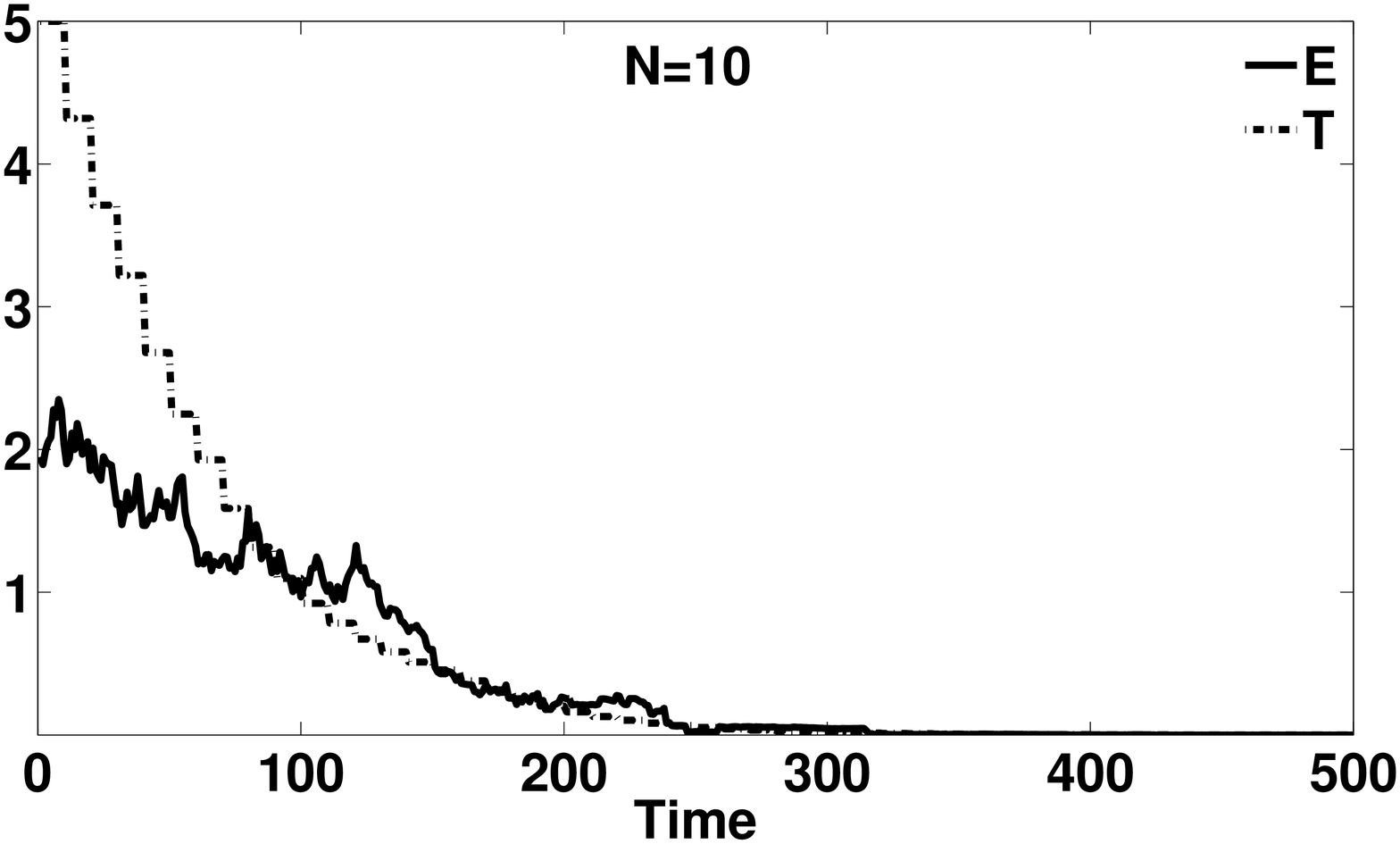}
&
\includegraphics[width=0.325 \textwidth, bb= -125 175 721 650]{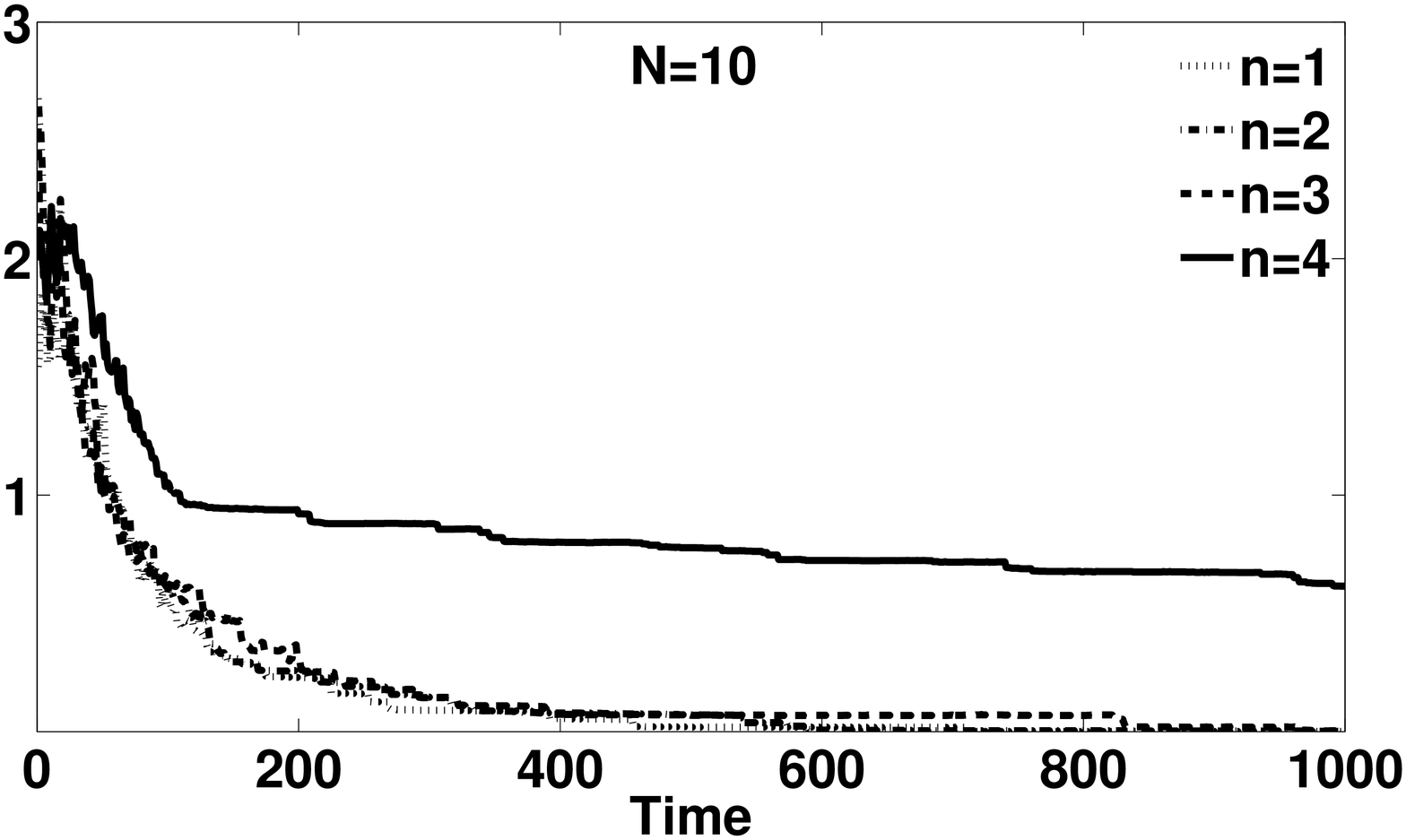}
&
\includegraphics[width=0.325 \textwidth, bb= -125 175 721 650]{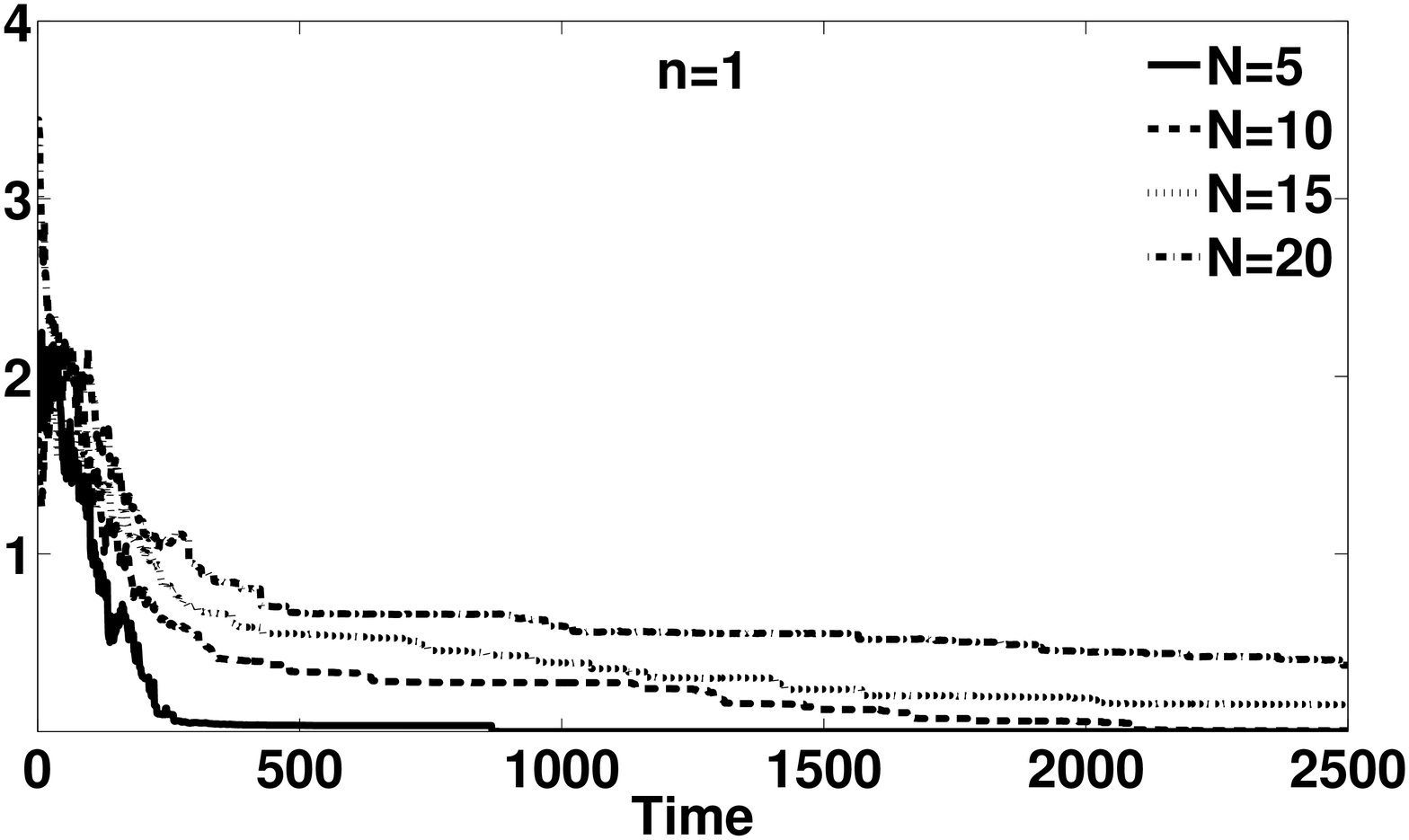}
\end{tabular}
\end{center}
\vspace{-0.3truecm} \caption{Mean error signal over 30 learning
sessions with Metropolis algorithm. Each time step corresponds to
the acceptance of just one local perturbation. a) Modulation of $T$
on $E$, fixed $N=10$, $n=1$; b) fixed $N=10$, $n$ variable; c) $N$
variable, fixed $n=1$.}
\end{figure}

Figure 4 shows the passage from an initial unconstrained dynamics to
an optimized one, ruled by the learning of a scheme. This transition
also corresponds to a passage from a $\zeta^{t_0}$ having one of
possible $\tau$ and $l$ (Fig.~4a), to a $\zeta^{t_{meas}}$ having
$\tau$ and $l$ respectively equal to zero and one (Fig.~4b).
\begin{figure} [h]
\begin{center}
\begin{tabular}{cc}
a&b \\ \\ \hspace{-.7truecm}
\includegraphics[width=0.51 \textwidth, bb= -125 175 721 650]{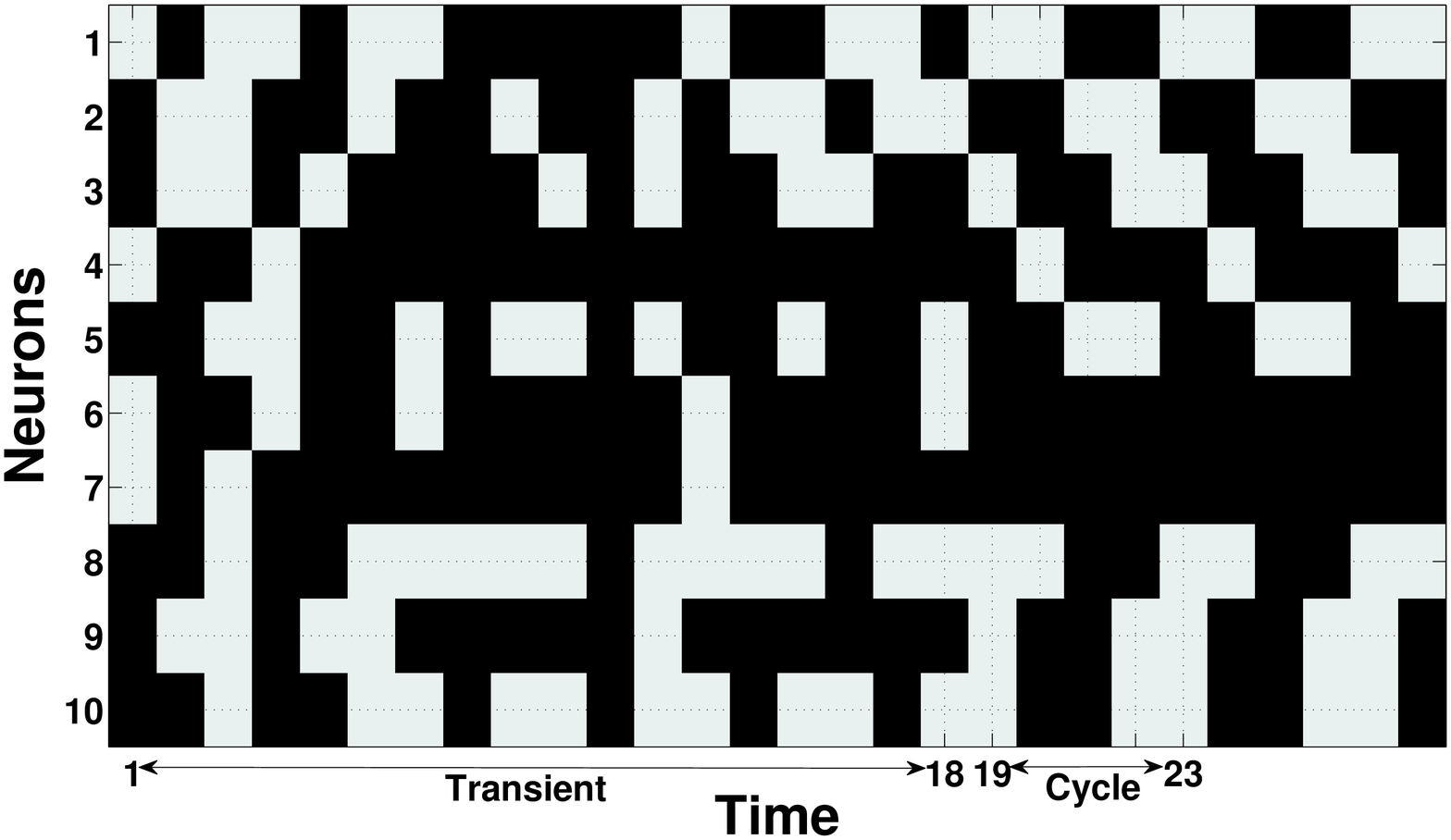}
&
\includegraphics[width=0.505 \textwidth, bb= -125 175 721 650]{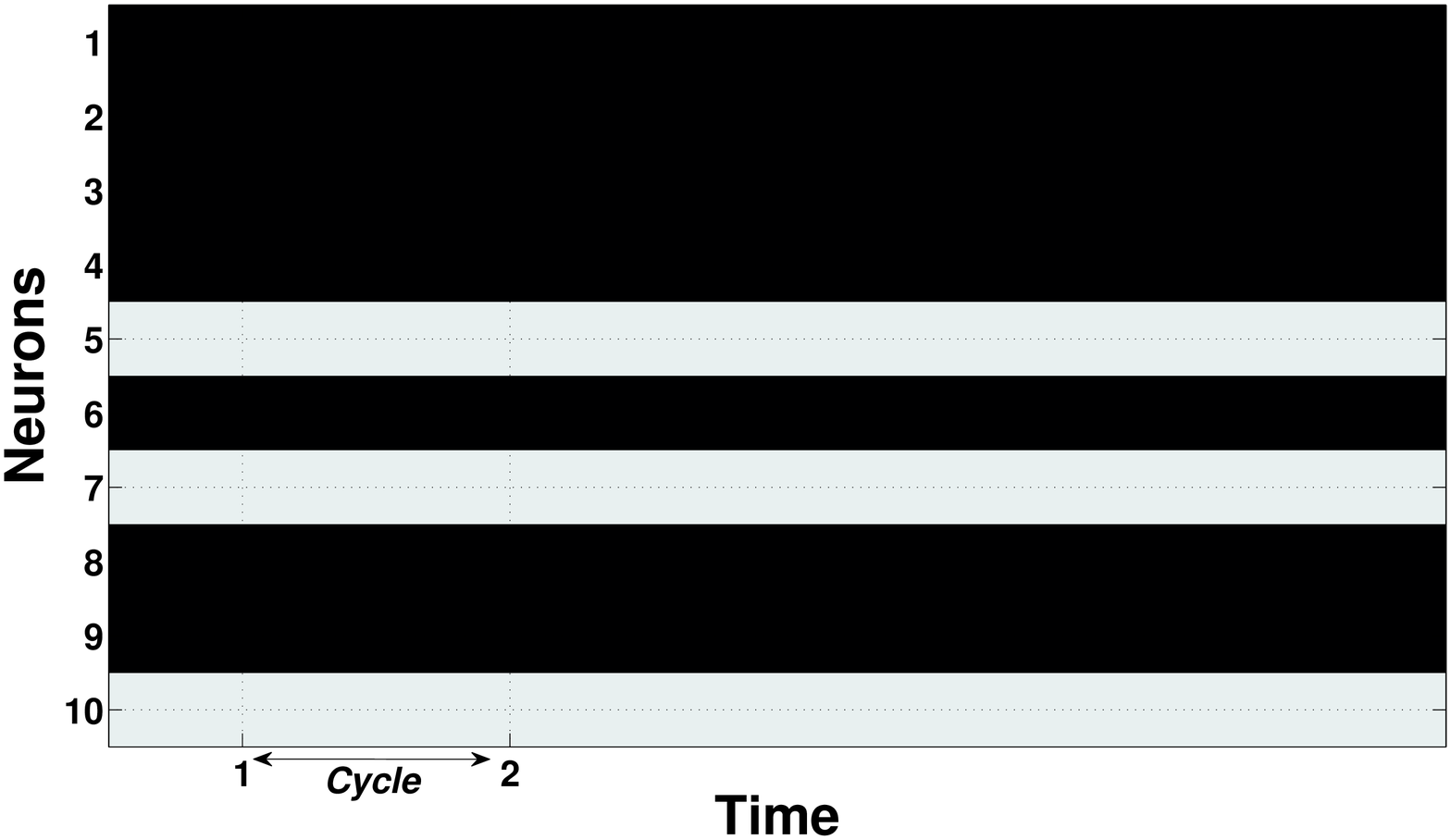}
\end{tabular}
\end{center}
\vspace{-0.3truecm} \caption{Dynamics transition. Neurons activation
(dark/black = +1; light/white = -1) as function of time. Each time
step is just one application of (\ref{dynamics}).}
\end{figure}

\section{Testing the model}
In this section we introduce a simplified version of
IGT\footnote{The original IGT is a psychological task used into
larger test batteries to study qualitatively behavior of normal
subjects during simulated real-life decision making. In
neuropsychological practice, it is administered to pathological
gamblers.} in order to test qualitatively the predictions of the
model.

The task consists of trials where a subject must select a card,
reporting both a term of winning and and a term of loss, from 4
decks (A', B', C' and D', respectively) of 60 cards. Main goal of
the subject is to maximize its budget after a 100 trials session.
The temporal series of decks are different; decks A' and B' promise
strong winning in the short period but stronger losses in the long
period, while C' and D', promising small winnings but also smaller
losses, assure a better budget at the end of the task.

For our purpose, the task to perform by our network take in
consideration only the two native decks B' (min = -2330; max = 170;
mean = -62.5) and D' (min = -310; max = 95; mean = -31.25), which
length is maintained to 60. While the choice of the first card is
random, the following ones are given by the output of the optimized
network, composed by 5 neurons, one of which is the input and one
the output. At each trial $Tr$, the network computes all the $Tr -
1$ previous trials as examples into a simulated annealing
optimization step, using the previous computed choices as computed
outputs $S^{out}$ and the terms of winning or loss in the
corresponding temporal series as wanted outputs $\xi^{out}$. This
procedure assumes in psychological terms an infinite working memory,
and explicitly makes use of the heuristic of availability; in
absence of relevant information about the covered cards, humans tend
to use available information stored from past trials. Once
optimized, the output computed by the asymptotical configuration
will become the choice of the new trial, and its value is registered
for future choices.

\begin{figure}
\begin{center}
\includegraphics[width=0.7 \textwidth, bb= -87 190 700 645]{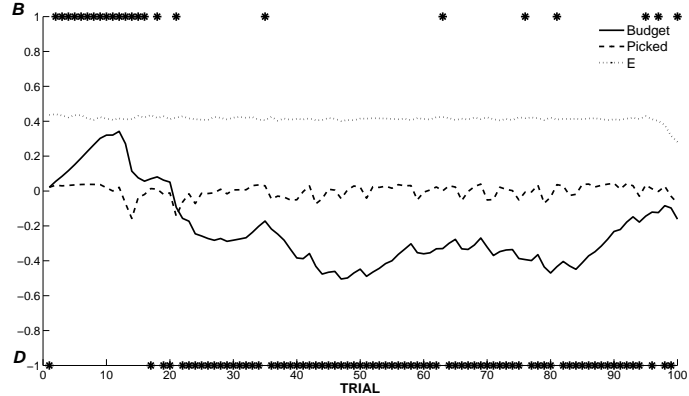}
\end{center}
\caption{Behavior of the network over 30 runs. Each time step is
just one trial. Red starred dots are most frequent choice of B',
green for D'.}
\end{figure}

Figure 5 shows typical behavior of the network while performing the
simplified IGT. The winnings early promised by B' prematurely
influences the network response in favor of deck B' but, after the
first severe losses, the functional $E$ associated to B' becomes too
large and the computed output switch in favor of D'. From this
moment on, almost all the perturbations to the asymptotical
configuration $\zeta^{t_{meas}}$ are rejected by the simulated
annealing and the network quickly produce its choice trial by trial.
It is interesting to point out that the transition from a regime
distinguished by choices in B' to one distinguished by choice in D'
happens approximately at the same time of humans \cite{12,13},
implying that the time scales in which the mental model becomes
effective are comparable between humans and our model.

Moreover after the first great loss given by B', network tends to
persevere with choices in the same deck, as both normal subjects and
pathological gamblers. This strange behavior can be interpreted from
a psychological point of view as a persistence of the use of the
heuristics of anchoring and adjustment during earlier trials after
the loss, while in our network is due to the fact that the error
quotas related to B' and D' becomes comparable.

\section{Conclusion and future perspectives}
We have presented preliminary results of application of a Boolean
model of neural network to relevant cognitive strategies involved in
decision making tasks. At moment only mental schemes have been
studied. The choice of a such formalization is due to the
possibilities that Boolean neural networks offers in terms of
robustness, ease of simulation and easy generation of samples for
data fitting.

The model appears to capture the most relevant psychological
knowledge regarding the domain of application. By shifting from an
unstructured slow attractor neural network to a quicker forward-only
one, it hold in respect of learning studies about task complexity
and the number of employed cognitive resources. As defined, mental
schemes become fast and adaptive cognitive strategies of behavioral
response.

Regarding the model fitting, the behavior of our network on the
simplified version of the IGT produces results qualitatively
comparable with those of humans.

Future studies will be targeted to include into the model aspects
regarding probabilistic and hypothetical-deductive reasoning, while
future applications will take in consideration pathological
gambling.

\appendix
\section{Implementation algorithm}
The model above described has been implemented with Matlab R2007a.
For sakes of space, the codes and the temporal series employed for
the model fitting are available by directly contacting the authors.


\begin{thebibliography}{a}

\bibitem{1} Abelson, R.P.,
    Psychological status of the script concept, American Psychologist, 36, 715-729 (1981).

\bibitem{2} Schank, R.C. and Abelson, R.P.,
    Scripts, plans, goals and understanding. Lawrence Erlbaum Associates
    Inc., Hillsdale, N.J. (1977).

\bibitem{3} Piaget, J.,
    Piaget's theory. In Carmichael's manual of child psychology, Vol. 1. J. Mussen (Ed.),  New York: Basic Books (1970).

\bibitem{4} Schank, R.C.
    Conceptual dependency: a theory of natural language understanding, Cognitive Psychology, 3, 552-631 (1972).

\bibitem{5} Wason, P.C.,
    Reasoning. In New horizons in psychology. Foss, B. M. (Ed.), Penguin, Harmondsworth (1966).

\bibitem{6} Jhonson-Laird, P.N.,
    Models of deduction. In Reasoning: representation and process in childreen and adults. Falmagne, R.J. (Ed.), Lawrence Erlbaum Associates
    Inc., Hillsdale, N.J. (1975).

\bibitem{7} Jhonson-Laird, P.N.,
    Mental models. Cambridge University Press, Cambridge (1983).

\bibitem{8} Jhonson-Laird, P.N. and Byrne, R.M.J.,
    Deduction. Lawrence Erlbaum Associates Ltd., London (1990).

\bibitem{9} McCulloch, W. and Pitts, W.,
    A logical calculus of the ideas immanent in nervous activity, Bulletin of Mathematical Biophysics, 7, 115-133 (1943).

\bibitem{10} Tversky, A. and Kahneman, D.,
    Judgment under Uncertainty: Heuristics and Biases, Science, New Series, 185, 4157, 1124-1131 (1974).

\bibitem{11} Bechara, A., Damasio, A. R., Damasio, H., and Anderson, S. W.,
    Insensitivity to future consequences following damage to human prefrontal cortex, , Cognition, 50, 7-15 (1994).

\bibitem{12} Bechara A., Damasio H., Tranel D., and Damasio A.R.,
    Deciding advantageously before knowing the advantageous strategy, Science, 275,
    1293-1295 (1997).

\bibitem{13} Bechara, A., Damasio, H., Tranel, D., and Damasio, A.R.,
    The Iowa Gambling Task and the somatic marker hypothesis: some questions and answers, Trends in Cognitive Sciences, 9, 4, 159-162 (2005).

\bibitem{14} http://www.breezetree.com/downloads/flow-chart-symbols.pdf

\end{thebibliography}
\end{document}